%% file: tnse.tex
\documentclass[10pt,journal,compsoc]{IEEEtran}
\ifCLASSOPTIONcompsoc
  \usepackage[nocompress]{cite}
\else
  \usepackage{cite}
\fi
\ifCLASSINFOpdf
\else
\fi
\usepackage{url} 
\usepackage{wrapfig}
\usepackage{graphicx}
\usepackage{textcomp}
\usepackage{xcolor}
\usepackage{mathtools}
\usepackage{threeparttable}
\usepackage{autonum}
\usepackage{tabularx}
\newcolumntype{R}{>{\raggedleft\arraybackslash}X}
\usepackage{enumitem}
\usepackage{microtype}
\usepackage{graphicx}
\usepackage{subcaption}
\usepackage{booktabs} 

\usepackage{tabularx}
\usepackage{caption}
\usepackage{amsmath}
\DeclareMathOperator{\msg}{MESSAGE}
\DeclareMathOperator{\agg}{AGGREGATE}
\DeclareMathOperator{\upd}{UPDATE}
\DeclareMathOperator{\ef}{F1}

\usepackage[backgroundcolor=yellow,textwidth=2cm]{todonotes}

\newcommand{\X}[1]{\underline{#1}}
\newcommand{\B}[1]{%
    \pdfliteral direct {2 Tr 0.3 w} 
     #1%
    \pdfliteral direct {0 Tr 0 w}%
}

\begin{document}
%
\title{Schema-Aware Deep Graph Convolutional Networks for Heterogeneous Graphs}
%
%
%
%

\author{Saurav~Manchanda,~\IEEEmembership{Member,~IEEE,}
        Da~Zheng,~\IEEEmembership{Member,~IEEE,}
        and~George~Karypis,~\IEEEmembership{Fellow,~IEEE}
\IEEEcompsocitemizethanks{\IEEEcompsocthanksitem This work was done when S. Manchanda was with the AWS AI Labs, Palo Alto.\protect\\
E-mail: manch043@umn.edu
\IEEEcompsocthanksitem D. Zheng and G. Karypis are with AWS AI Labs, Palo Alto.}
\thanks{Manuscript received April 19, 2005; revised August 26, 2015.}}

%
%

\markboth{Journal of \LaTeX\ Class Files,~Vol.~14, No.~8, August~2015}%
{Shell \MakeLowercase{\textit{et al.}}: Bare Demo of IEEEtran.cls for Computer Society Journals}
%



\IEEEtitleabstractindextext{%
\begin{abstract}
\input{abstract}
\end{abstract}

}

\maketitle

\IEEEdisplaynontitleabstractindextext

%
\IEEEpeerreviewmaketitle

\IEEEraisesectionheading{\section{Introduction}\label{sec:introduction}}

%
%
%
%
\input{introduction}
\input{notation}
\input{related}
\input{dhgcn}
\input{methodology}
\input{results}
\input{conclusion}



\ifCLASSOPTIONcaptionsoff
  \newpage
\fi



%
\bibliographystyle{IEEEtran}
\bibliography{refs}

%

\begin{IEEEbiography}[{\includegraphics[width=1in,height=1.25in,clip,keepaspectratio]{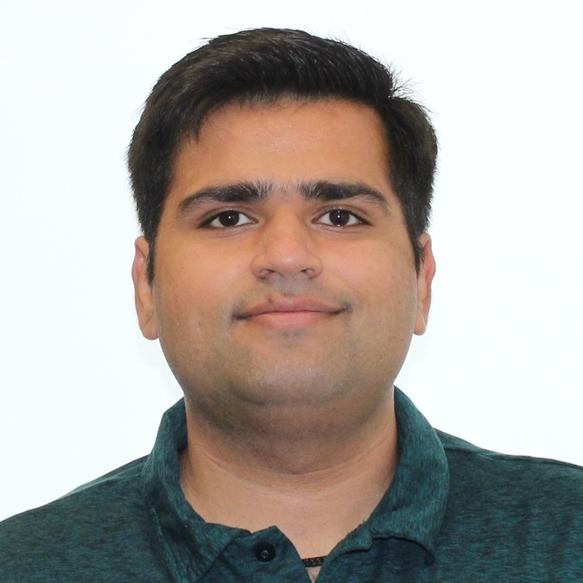}}]{Saurav Manchanda}
is affiliated with Instacart, where he works on applied machine learning. His primary interests involve data mining, information retrieval, and graph analytics. He got a PhD in computer science at the University of Minnesota, Twin Cities, USA in 2020 and B.Tech. (Bachelor of Technology) from Indian Institute of Technology (IIT), Kharagpur, India in 2015.
\end{IEEEbiography}

\begin{IEEEbiography}[{\includegraphics[width=1in,height=1.25in,clip,keepaspectratio]{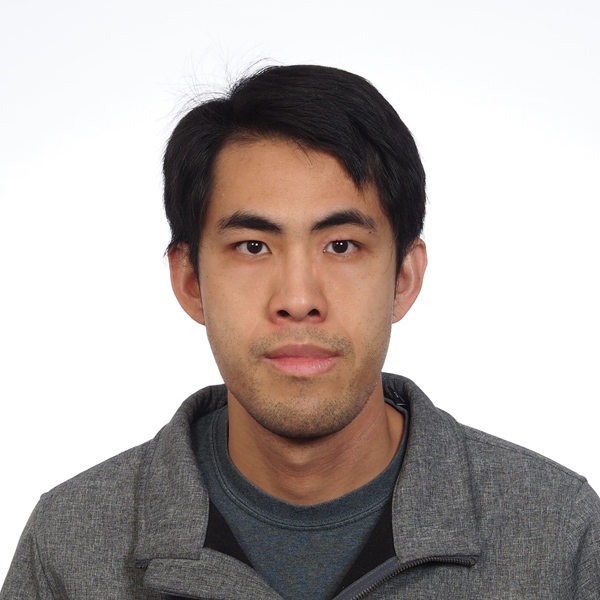}}]{Da Zheng}
is a senior applied scientist at AWS AI, where he develops deep learning frameworks including MXNet, DGL (Deep Graph Library) and DGL-KE. His research interest includes high-performance computing, scalable machine learning systems and data mining. He got a PhD in computer science at the Johns Hopkins University, Baltimore, USA in 2017. He received his master of science from École polytechnique fédérale de Lausanne, Lausanne, Switzerland in 2009 and bachelor of science from Zhejiang University, Hangzhou, China in 2006.
\end{IEEEbiography}


\begin{IEEEbiography}[{\includegraphics[width=1in,height=1.25in,clip,keepaspectratio]{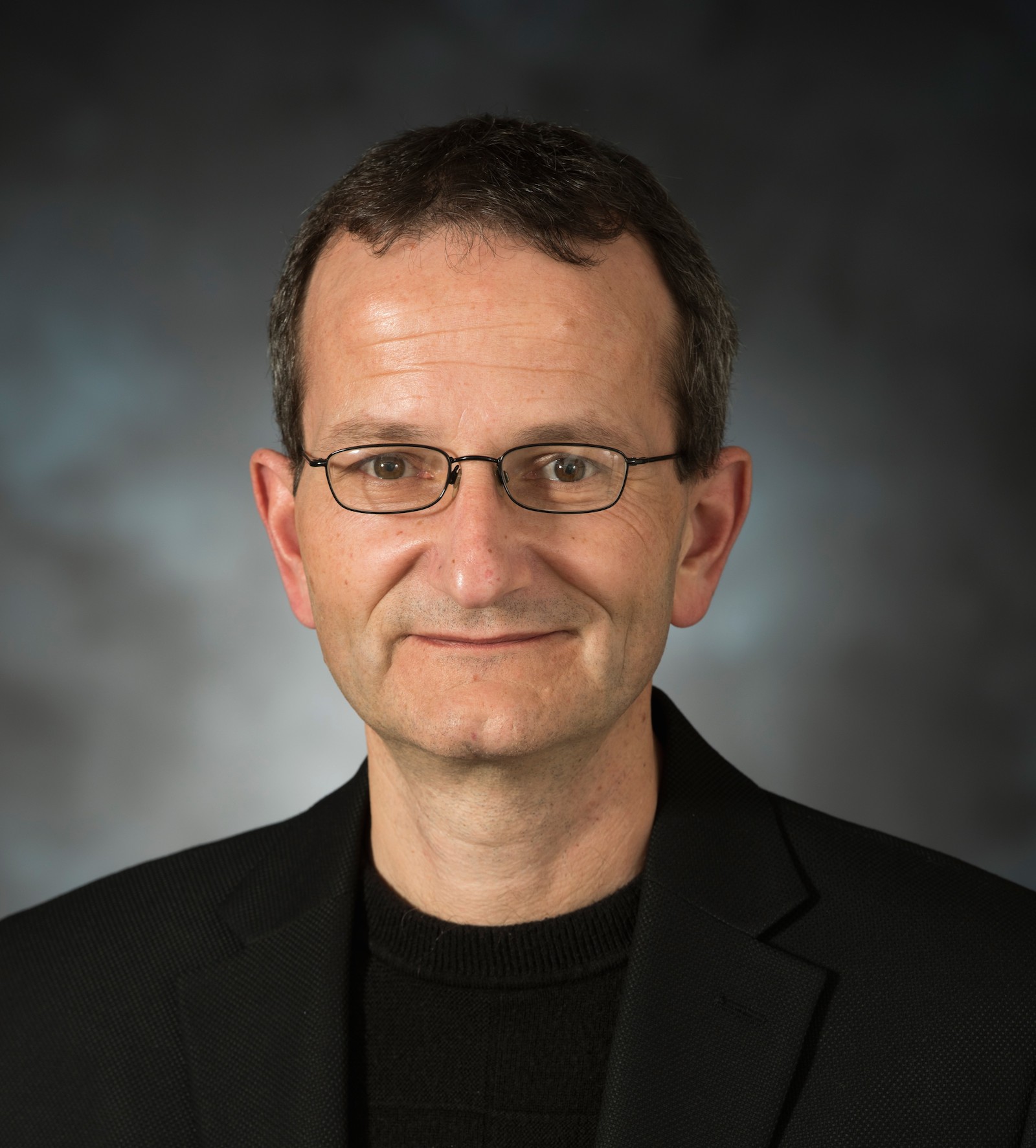}}]{George Karypis}
is a Senior Principal Scientist at Amazon, a Distinguished McKnight University Professor and an ADC Chair of Digital Technology at the Department of Computer Science \& Engineering at the University of Minnesota, Twin Cities. His research interests span the areas of data mining, high performance computing, information retrieval, collaborative filtering, bioinformatics, cheminformatics, and scientific computing. He has coauthored over 300 papers on these topics and two books (``Introduction to Protein Structure Prediction: Methods and Algorithms'' (Wiley, 2010) and ``Introduction to Parallel Computing'' (Publ. Addison Wesley, 2003, 2nd edition)). In addition, he is serving on the program committees of many conferences and workshops on these topics, and on the editorial boards of the ACM Transactions on Knowledge Discovery from Data, Data Mining and Knowledge Discovery, Social Network Analysis and Data Mining Journal, International Journal of Data Mining and Bioinformatics, the journal on Current Proteomics, Advances in Bioinformatics, and Biomedicine and Biotechnology. He is a Fellow of the IEEE.
\end{IEEEbiography}




\end{document}

%% file: abstract.tex
Graph convolutional network (GCN) based approaches have achieved significant progress for solving complex, graph-structured problems. GCNs incorporate the graph structure information and the node (or edge) features through message passing and computes `deep' node representations. Despite significant progress in the field, designing GCN architectures for heterogeneous graphs still remains an open challenge. Due to the schema of a heterogeneous graph, useful information may reside multiple hops away. A key question is how to perform message passing to incorporate information of neighbors multiple hops away while avoiding the well-known over-smoothing problem in GCNs. To address this question, we propose our GCN framework \emph{Deep Heterogeneous Graph Convolutional Network (DHGCN)}, which takes advantage of the schema of a heterogeneous graph and uses a hierarchical approach to effectively utilize information many hops away. It first computes representations of the target nodes based on their \textit{schema-derived ego-network} (SEN). It then links the nodes of the same type with various pre-defined metapaths and performs message passing along these links to compute final node representations. Our design choices naturally capture the way a heterogeneous graph is generated from the schema. The experimental results on real and synthetic datasets corroborate the design choice and illustrate the performance gains relative to competing alternatives.

%% file: introduction.tex
Many real-world problems are naturally represented as graphs. Inspired by the success of Convolutional Neural Networks (CNNs)~\cite{krizhevsky2012imagenet} in computer vision, Graph Convolution Networks (GCNs)~\cite{hamilton2017inductive, zhang2019star, li2015gated, kipf2016semi, zhang2018gaan, velickovic2018graph} defined on graphs have emerged as one of the most powerful tools for solving graph problems. 
GCNs apply a convolutional layer in the neighborhood of a node to generate node representation, where the neighbors generally correspond to directly connected nodes. By stacking multiple such convolutional layers, GCNs compute \emph{deep} node representations by using information from nodes multiple hops away.


While there is significant progress in the field, designing GCN architectures for heterogeneous graphs still remains an open challenge. Heterogeneous graphs consist of multiple node types and relation types and can be viewed as graphs with pre-defined network schema, where some nodes are entity nodes and other nodes are attribute nodes describing these entity nodes. The same entity can be represented using different subgraphs of attributes. For example, Figure~\ref{fig:schema} shows two network schemas that model the same information. Depending on the schema, some attribute nodes may be far away from entity nodes. This gives rise to the question: can we design a GCN architecture for heterogeneous graphs such that it can effectively get information from attribute nodes many hops away as well as entity nodes of the same type.

Some heterogeneous GNN models, such as Relational Graph Convolution Network (RGCN)~\cite{schlichtkrull2018modeling}, performs message passing with direct neighbors. One may argue that in order to get information from nodes multiple hops away we just need to put more RGCN layers. However, there are two limitations with such an approach: (i) Since important information can be present many hops away, the number of RGCN layers can grow very large. However, by increasing the number of RGCN layers, we also bring irrelevant information to the model.
(ii) GNNs with multiple layers suffer from an over-smoothing problem~\cite{li2018deeper, luan2019break, wu2019simplifying}, in which all nodes will converge to the same representation. 
Another line of research on heterogeneous graphs takes advantage of
metapaths~\cite{fu2020magnn, wang2019heterogeneous}. By using metapaths, these models can reach
the information on the nodes multiple hops away. For example, HAN~\cite{wang2019heterogeneous} performs
message passing between nodes of the same type
that are linked with pre-defined metapaths. However, HAN only uses the information on the endpoint nodes
of the metapath and ignores all nodes in the ego-network of these endpoint nodes as well as the nodes along with
the metapath. Thus, HAN can potentially miss much information in a heterogeneous graph.
MAGNN~\cite{fu2020magnn} improves HAN by incorporating the information of the nodes along the metapath
but still misses the information in the ego-network of the endpoint nodes.

%

We introduce the \emph{Deep Heterogeneous Graph Convolutional Network} (DHGCN), a purposefully designed class of GCN models for heterogeneous graphs. DHGCN uses a two-step schema-aware hierarchical approach. Without loss of generality, assume that we are interested in making predictions for a node of node type $\mathcal{A}$. In the first step, DHGCN aggregates information from the immediate ego-network that is derived from the schema. Second, DHGCN aggregates information from the neighbors of type $\mathcal{A}$ using metapath-based convolutions, i.e., graph convolutions on the metapath-guided neighbors.
%
%
In a heterogeneous graph, we believe that our proposed framework is the right way to aggregate the information because it captures the way the graph was generated using the schema. 

We present extensive experiments using various real and synthetic datasets to explore DHGCN's design space and assess its performance. Though DHGCN can be used for many problems that involve learning over graphs, we evaluate our approaches on node classification. We show that as the complexity of the graph increases, DHGCN outperforms the competing approaches. On various datasets, DHGCN achieves up to $10.4\%$ improvement over the competing approaches. 

The remainder of the paper is organized as follows.  Section \ref{sec:definitions} corresponds to the definitions, notations and the background used in the paper. The paper discusses the proposed methods in Section \ref{sec:proposed} followed by the experiments in Section \ref{sec:experiments}. Section \ref{sec:results} discusses the results.  Finally, Section \ref{sec:conclusion} provides some concluding remarks.

%% file: notation.tex
\begin{figure}[!t]%
    \centering
    \subfloat[]{\includegraphics[width=0.35\columnwidth]{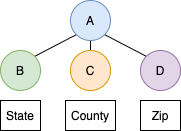} \label{fig:schema:1}}\hfil%
    \subfloat[]{\includegraphics[width=0.40\columnwidth]{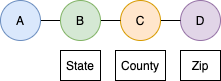} \label{fig:schema:2}}%
    \caption{The same entity `address' can be represented in multiple ways, depending upon the design of the network schema.}%
    \label{fig:schema}%
\end{figure}


\section{Notation and Background}\label{sec:definitions}
\begin{table}[!t]
\small
\centering
  \caption{Notation used throughout the paper.}
  \begin{tabularx}{\columnwidth}{lX}
    \hline
Symbol   & Description \\ \hline
$\mathcal{G}$    & A heterogeneous graph. \\
$\mathcal{V}$    & Set of nodes in the graph $\mathcal{G}$. \\
$\mathcal{E}$    & Set of edges in the graph $\mathcal{G}$. \\
$\mathcal{A}$    & Set of node types in the graph $\mathcal{G}$. \\
$\mathcal{R}$    & Set of edge types in the graph $\mathcal{G}$. \\
$\tau(v)$    & Type mapping function $\tau(v) : \mathcal{V} \rightarrow \mathcal{A}$, that maps a node to its corresponding type in the graph $\mathcal{G}$. \\
$\phi(e)$    & Type mapping function $\tau(e) : \mathcal{E} \rightarrow \mathcal{R}$, that maps an edge to its corresponding type in the graph $\mathcal{G}$. \\
$\psi(p)$    & Type mapping function that maps a metapath instance to its corresponding metapath in the graph $\mathcal{G}$. \\
$P$    & A metapath $P$ on the heterogeneous graph $\mathcal{G}$ is defined as a path on $\mathcal{G}$'s network schema, of the form $A_1 \xrightarrow{R_{1,2}} A_2 \xrightarrow{R_{2,3}} \dots \xrightarrow{R_{n-1,n}} A_{n}$. \\
$E_a$    & The \emph{schema-derived ego-network} (SEN) of node $a$ of graph $\mathcal{G}$  \\
$z_a$    & Aggregated information of the SEN $E_a$  \\
$N^l(v)$    & A function that defines the neighbor nodes of a node $v \in \mathcal{V}$ for $l$th layer in a GCN framework.  \\
$N_m^l(v)$    & A function that defines the metapath $m$ guided neighbor nodes of a node $v \in \mathcal{V}$ for $l$th layer in a GCN framework.  \\
$q^{l}_{m,\hat{v}\rightarrow v}$    & Path-aware attention-weight for the message from node $\hat{v}$ to the node $v$ through an instance of the metapath $m$, for layer $l$ of metapath-based convolutions.\\
$u^{l}_{m,\hat{v}\rightarrow v}$    & Unnormalized attention-weight corresponding to $q^{l}_{m,\hat{v}\rightarrow v}$, estimated as a function of source node features, destination node features, in addition to metapath-features.\\
$W_a^b$    & Projection matrix, the subscripts and superscripts are used to differentiate between the different projection matrices.\\
$f_v$    & Features associated with a node $v$.\\
$h_v$    & Representation of a node $v$.\\
$h_v^l$    & Representation of a node $v$ at layer $l$.\\
$\sigma$    & Non-linear operation.\\

\hline
\end{tabularx}
  \label{tab:notation}
\end{table}
\subsection{Heterogeneous graphs}
A heterogeneous graph is a directed graph $\mathcal{G}(\mathcal{V}, \mathcal{E}, \mathcal{A}, \mathcal{R})$, where each node $v \in \mathcal{V}$ and each edge $e \in \mathcal{E}$ are associated with their type mapping functions $\tau(v) : \mathcal{V} \rightarrow \mathcal{A}$ and $\phi(e) : \mathcal{E} \rightarrow \mathcal{R}$, respectively. A metapath $P$ on the heterogeneous graph $\mathcal{G}$ is defined as a path on $\mathcal{G}$'s network schema, of the form $A_1 \xrightarrow{R_{1,2}} A_2 \xrightarrow{R_{2,3}} \dots \xrightarrow{R_{n-1,n}} A_{n}$, 
which describes a composite relation $R = R_{1,2} \circ R_{2,3} \circ \dots \circ R_{n-1,n}$ between node types $A_1$ and $A_n$. Given a metapath $P$ of a heterogeneous graph, a metapath instance $p$ of $P$ is defined as a node sequence in the graph following the schema defined by $P$.

\subsection{Metapath-guided neighbors}

Given a metapath $P$, where $P$ is of the form $A_1 \xrightarrow{R_{1,2}} A_2 \xrightarrow{R_{2,3}} \dots \xrightarrow{R_{n-1,n}} A_{n}$,  the metapath-guided
neighbors of a node $v\in A_1$ is defined as the set of all the nodes, belonging to the same type $ A_n$, that are reachable from $v$, following the metapath $P$.

When the metapath $P$ of the form $A_1 \xrightarrow{R_{1,2}} A_2 \xrightarrow{R_{2,3}} \dots \xrightarrow{R_{n-1,n}} A_{n}$ has $A_{1} = A_{n}$, the metapath-guided neighbors are of the same type $A_{1}$ (or $A_{n}$). We call these metapath-guided neighbors as the same-type metapath-guided neighbors, and these neighbors carry a special semantic meaning. 
For example, in a scholar network with authors, papers, and venues, Author-Paper-Author (APA) and Author-Paper-Venue-Paper-Author
(APVPA) are symmetric-metapaths describing two different relations among
authors. The APA metapath associates two co-authors, while the
APVPA metapath associates two authors who published papers in
the same venue.

\subsection{Message passing}
Graph neural network models can be formulated in the form of message passing
\cite{gilmer2017neural}, in which each node of a graph broadcasts messages to its
neighbors and compute its node representation with its own features as well as
the messages from its neighbors. A general form of the message passing mechanism has four functions:
\begin{description}[style=unboxed,leftmargin=0cm]
    \item[Neighbor identification: ]A function $N^l(v)$ that defines
        neighbor nodes of a node $v \in \mathcal{V}$ for $l$th layer. 
    \item[Message construction: ]A function that maps the representation $h_{\hat{v}}$ of a neighbor node $\hat{v} \in N^l(v)$ of a target node $v$ to form a message:
    \begin{equation}
        m^l_{\hat{v}\rightarrow v} = \msg^l (h^{l-1}_{\hat{v}}, h^{l-1}_{v}, c(\hat{v}, v)),
    \end{equation}
    where $c(\hat{v}, v)$ is the connectivity information between $\hat{v}$ and $v$. For example, when $N^l(v)$ corresponds to directly connected nodes of $v$, $c(\hat{v}, v)$ can be the features on the edge connecting $v$ and $\hat{v}$, i.e., $e^l_{\hat{v}\rightarrow v}$.
    \item[Message aggregation: ]A function that gathers the messages from a
node $v$'s neighbors:
    \begin{equation}
        m^l_{v} = \agg^l(\{m_{\hat{v}\rightarrow v} |\hat{v} \in N^l(v)\})
    \end{equation}
    \item[Node update: ]A function that updates the node $v$'s representation by the aggregated message $m^l_{v}$ and its representation at previous layer $h^{l-1}_{v}$
    \begin{equation}
        h^l_{v} = \upd^l(m^l_{v}, h^{l-1}_{v})
    \end{equation}
\end{description}
\noindent All of these functions can be defined with neural networks and can be learned
for a downstream task.

%% file: related.tex
\section{Related work}
Many research works have been conducted to develop machine learning models
for graphs. Earlier works \cite{Perozzi_2014, grover2016node2vec, Tang_2015, metapath2vec}
developed unsupervised models to learn representations of nodes in a graph.
The representations can be used in many downstream tasks. Although these unsupervised
models are powerful, they have some limitations. First, they are not able to incorporate
node or edge features in the learned node representations. Secondly,
the representations are learned with supervisions that are different from
the downstream tasks.

Graph neural networks arose in popularity in the past few years. They apply deep neural networks on
graph data and can address the limitations in the earlier representation learning
methods. GNN models apply message passing on graphs and compute node representations
with input node/edge features and graph structures.
These models can be trained with any downstream
tasks to learn node representations specifically for these tasks and achieve better
performance.


The earlier GNN models \cite{hamilton2017inductive, li2015gated, velickovic2018graph, kipf2016semi}
apply message passing on homogeneous graphs.
The initial GCN network, proposed by~\cite{kipf2016semi}, uses all one-hop direct nodes to $v$, in addition to $v$ (i.e., self-edge) as $N^l(v)$. The $\msg$ function is modeled as a projection of $h^{l-1}_{\hat{v}}$ through a weight matrix $W^l$. The $\agg$ function is modeled by taking sum of the incoming messages and normalizing it. The $\upd$ function simply returns the aggregated message $m^l_{v}$ followed by a non-linearity $\sigma$. Formally, 
\begin{equation}
    h^l_{v} = \sigma\left(\sum_{\hat{v} \in N^l(v)}\frac{1}{d_{v\hat{v}}}W^lh^{l-1}_{\hat{v}}\right),
\end{equation}
where $d_{v\hat{v}} = \sqrt{|N(v)||N(\hat{v})|}$ is a normalization constant based on graph structure. Graph attention network (GAT)~\cite{velickovic2018graph} extends GCN by introducing the attention mechanism as a substitute for the statically normalized convolution operation. The attention weights are used to do a weighted combination of the messages in the $\agg$ function, as compared to the vanilla sum used in GCN.
We can apply message passing on a graph multiple times to have a node to gather
information from nodes multiple hops away, instead of just direct neighbors.

The message passing formulation of graph neural networks is extended to heterogeneous
graphs.
Relational Graph Convolution Network (RGCN)~\cite{schlichtkrull2018modeling} uses
the type-specific projection matrix for message construction, i.e., 
\begin{equation}
    h^l_{v} = \sigma\left(\sum_{\hat{v} \in N^l(v)}\frac{1}{d_{v\hat{v}}}W^l_{\phi(e(\hat{v}, v))}h^{l-1}_{\hat{v}}\right),
\end{equation}
where $W^l_{\phi(e(\hat{v}, v))}$ is a type-specific projection matrix and $\phi(e(\hat{v}, v))$ gives the edge-type (relation) between the nodes $\hat{v}$ and $v$.
Heterogeneous graph transformer \cite{hu2020heterogeneous} applies the attention
mechanism to heterogeneous graphs. The attention score on each edge is parametrized
by the relation type.

A major problem of GNN models is that they suffer from an over-smoothing problem
when multiple GNN layers are deployed to access data from neighbors multiple hops
away \cite{chen2019measuring}.
The problem is even more pronounced in the case of heterogeneous graphs, where the nodes of interest can be present at arbitrary depth, depending upon the underlying network schema. 

\begin{figure*}[!t]%
    \centering
    \hspace{10pt}
    \subfloat[HAN]{\includegraphics[width=0.40\columnwidth]{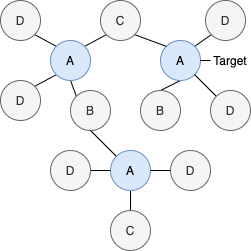} \label{fig:ego:han}}
    \hspace{50pt}
    \subfloat[MAGNN]{\includegraphics[width=0.40\columnwidth]{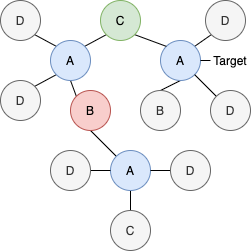} \label{fig:ego:magnn}}
    \hspace{50pt}
    \subfloat[DHGCN]{\includegraphics[width=0.40\columnwidth]{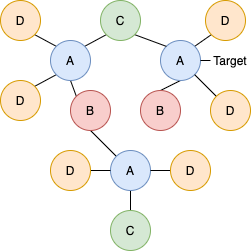} \label{fig:ego:proposed}}
    \caption{Ego-networks explored by various approaches for a given target-node belonging to type $A$ (greyed-out nodes are the ones not explored). HAN only explores the metapath-guided neighbors of the same-type $A$. MAGNN also explores other nodes that appear on the metapath $A...A$, thus not exploring the neighbors which are not part of the metapath. The proposed approaches explore the complete ego-network, without suffering through the oversmoothing problems faced by RGCN.}%
    \label{fig:ego}%
\end{figure*}

To address this challenge, several approaches have been developed to use the metapaths to guide neighbor selection, allowing the approaches to get information from distant nodes. The approaches that use metapath-guided neighbor selection include Heterogeneous Graph Attention Network (HAN)~\cite{wang2019heterogeneous} and Metapath Aggregated Graph Neural Network (MAGNN)~\cite{fu2020magnn}. These approaches differ in their neighbor-identification approach. While HAN only uses the metapath-guided neighbors as the neighbor nodes, MAGNN extends HAN to compute the representation with the information of the nodes that appear on the metapath. The ego-networks explored by HAN and MAGNN are shown in Figures~\ref{fig:ego:han} and~\ref{fig:ego:magnn}, respectively. Both HAN and MAGNN are prone to missing out some nodes, which might provide important signals for the task at hand. 

In this paper, we develop novel approaches to address the limitations of prior approaches.

%% file: dhgcn.tex
\section{Deep Heterogenous Graph Convolutional Networks}
\label{sec:proposed}
\begin{figure}[!t]%
    \centering
    \includegraphics[width=1.0\columnwidth]{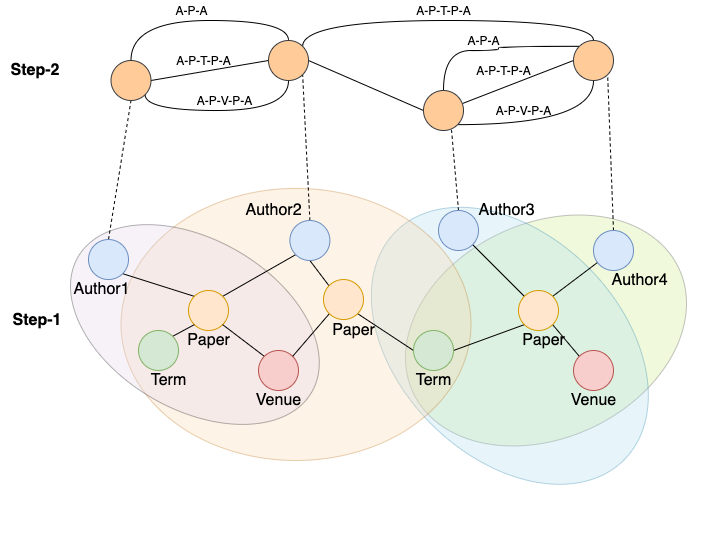}
    \caption{Overview of the proposed framework. The key characteristic of the framework is a hierarchical structure that first computes representations of the target entity-types (Author) based on their heterogeneous networks, i.e., Step-1 and then combine these using various metapaths into a homogeneous second-level graph, i.e., Step-2.}%
    \label{fig:overview}%
\end{figure}

%
We introduce the \emph{Deep Heterogeneous Graph Convolutional Network} (DHGCN), a purposefully designed class of GCN models for heterogeneous graphs. DHGCN uses a two-step schema-aware hierarchical approach. Without loss of generality, assume that we are interested in making predictions for a node of type $\mathcal{A}$. In the first step, DHGCN aggregates information from the immediate ego-network that is derived from the schema. Second, DHGCN aggregates information from the neighbors of type $\mathcal{A}$ using metapath-based convolutions, i.e., graph convolutions on the same-type metapath-guided neighbors. Figure~\ref{fig:overview} gives an overview of the two-step aggregation strategy of the proposed framework.

In the remaining part of this section, we discuss in detail about, (i) generating schema-derived ego-network, (ii) aggregating information from this schema-derived ego-network, and (iii) aggregating information from the same-type metapath-guided neighbors using metapath-based convolutions.

\subsection{Schema-derived ego-network}
We are interested in getting a representation (and hence making predictions) for node $a$ of type $\mathcal{A}$ in graph $\mathcal{G}$.
We define the \emph{schema-derived ego-network} (SEN) $E_a$ of node $a$ of graph $\mathcal{G}$ as a directed subgraph of $\mathcal{G}$, which contains all the nodes that describe $a$. $E_a$ can be derived from the graph schema automatically, but can also be customized by domain experts for the task at hand. In this paper, we take a general approach, and construct $E_a$ from the graph schema as follows: starting from $a$, we do a depth-first traversal of $G$, such that we do not visit the same edge type more than once in the same path (there is no loop). We denote the set of children (successors) of a node $v$ as $C(v)$. Note that SEN will contain a loop if and only if, the network schema of $\mathcal{G}$ contains a loop. 

\begin{figure}[!t]%
    \centering
    \subfloat[Hierarchical aggregation (HA)]{\includegraphics[width=0.35\columnwidth]{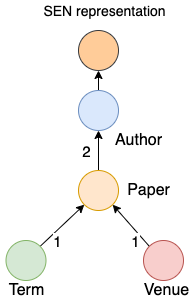} \label{fig:ha}}%
    \hfil
    \subfloat[Set aggregation (SA)]{\includegraphics[width=0.35\columnwidth]{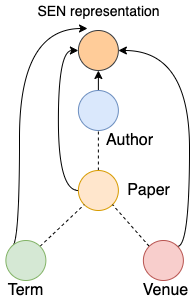} \label{fig:sa}}%
    \caption{Two approaches to aggregate information from the \emph{schema-derived ego-network} (SEN).}%
    \label{fig:step1}%
\end{figure}


\subsection{Step 1: Aggregating information from the \emph{schema-derived ego-network} (SEN)}
The first step involves aggregation of information from the SEN $E_a$ of $a$, as well as SENs of neighbors ${N^l(a)}$ of $a$. Lets denote the aggregated information of the  SEN $E_a$ of $a$ as $z_{a}$. This step can be modeled as any pooling operation over a subgraph. In this paper, we explore two approaches for this step:  \emph{Hierarchical-Aggregation} (HA) and \emph{Set-Aggregation} (SA). For HA, we aggregate the information of the SEN by propagating the information from the leaves towards the target node following a Tree-LSTM fashion, while for SA, we relax the hierarchy constraint, and model the node types in SEN as a set. Figure~\ref{fig:step1} illustrates the HA and SA on the scholar network with author, paper, venue and term node types.  We discuss these aggregation approaches in detail below:
\subsubsection{Hierarchical-Aggregation (HA)} For HA, we aggregate the information of the SEN by propagating the information from the leaves towards the target node following a Tree-LSTM fashion. Particularly, 
    \begin{equation}
        z_{a} = W_{\tau(a)}(f_a) + \sum_{v\in C(a)} W_{\phi(e(v,a))} (h_v),
    \end{equation}
    where $W_{\tau(a)}$ is a node-type specific projection matrix, $W_{\phi(e(v,a))}$ is an edge-type specific projection matrix, and $f_a$ are the features associated with the node $a$. $h_v$ is recursively defined as follows:
    \begin{equation}
        h_{v} = \sigma(W_{\tau(v)}(f_v) + \sum_{\hat{v}\in C(v)} W_{\phi(e(\hat{v},v))} (h_{\hat{v}})),
    \end{equation}
    where $\sigma$ is a non-linearity such as Tanh or ReLU.
    \subsubsection{Set-Aggregation (SA)}For SA, we relax the connectivity information in SEN, and estimate the SEN representation as a sum of the projected representations of the constituent nodes. Each metapath from the target node to the constituent nodes corresponds to a projection matrix. Particularly,
    \begin{equation}
        z_{a} = \sum_{v\in C(a)}\sum_{p\in M(a\rightarrow v)} W_{\psi(p)} (h_v),
    \end{equation}
    where $M$ is the set of metapath instances from $a$ to $v$, such that the metapath corresponding to the metapath instance does not contain a loop, and $\psi(p)$ is the mapping fuction which maps the metapath instance $p$ to its corresponding metapath. $W_{\psi(p)}$ is a metapath specific projection matrix.
    Though SA is not that expressive, as by relaxing the connectivity information, it essentially ignores the semantics that are encoded in the underlying schema, it is interesting in the view that it is invariant to schematic changes discussed before. A similar idea has also been leveraged by Network Schema Preserving Heterogeneous Information Network Embedding (NSHE)~\cite{zhao2020network}, which estimates node-representations that are aware of the subgraph imposed by the schema, and not just aware of the locally connected neighborhood.
    
    \subsection{Step 2: Aggregating information from the same-type metapath-guided neighbors}
    The same-type metapath-guided neighbor selection gives rise to another heterogeneous graph, which has only one node type (the target type) and the number of relations is equal to the number of metapaths used to guide neighbor selection. To model the aggregation step over this heterogeneous graph, we investigate a metapath-aware aggregation approach. Our aggregation approach has two components as follows:
    
\subsubsection{Step 2(a): Aggregating information individually for each metapath} The second step involves aggregation of information from the representations of the same-type metapath-guided neighbors ${N^l(a)}$ of $a$.
We propose a metapath-aware attention-based aggregation to aggregate the information from the metapath-neighbors. Particularly, we denote the aggregated information for a metapath $m$ at layer $l$ as $h^l_{m,v}$, and estimate it as
    
    \begin{equation}
        h^l_{m,v} = \sum_{\hat{v} \in N^l_m(v)} q^{l}_{m,\hat{v}\rightarrow v} h^{l-1}_{\hat{v}},
    \end{equation}
    where $N^l_m(v)$ is the set of metapath $m$ guided neighbors for node $v$ for $l$th layer, and $q^{l}_{m,\hat{v}\rightarrow v}$ are the path-aware attention-weights calculated as follows:
    
    \begin{equation}
        q^{l}_{m,\hat{v}\rightarrow v} = \frac{exp(u^{l}_{m,\hat{v}\rightarrow v})}{\sum_{\bar{v} \in N^l_m(v)} exp(u^{l}_{m,\bar{v}\rightarrow v})},
    \end{equation}
    where $u^{l}_{m,\hat{v}\rightarrow v}$ is further a function of source node features, destination node features, in addition to metapath-features.  Specifically, given the metapath instance of the metapath $m$ denoted by $\hat{v} \rightarrow \bar{v}_1 \rightarrow \bar{v}_2 \dots \bar{v}_{|p|-2} \rightarrow v$, we estimate $u^{l}_{m,\hat{v}\rightarrow v}$ as follows:
    \begin{equation}
        u^{l}_{m,\hat{v}\rightarrow v} = \sigma(\hat{W}^l_{m,u} (h^{l-1}_{\hat{v}} || (||_{i=1}^{|p|-2} f_{\bar{v}_{i}})|| h^{l-1}_{v} )),
    \end{equation}
    where $||$ is the concatenation operator and $\hat{W}^l_{m,u}$ is a projection matrix. Without loss of generality, we can assume that the size of all the features-vectors of the nodes as well as the latent representations is same and equals $d$. This makes the size of the projection matrix $\hat{W}^l_{m,u}$ as $(|p|*d, 1)$.
    Particularly, we concatenate the $l-1$ layer representations of the source and destination nodes, along with the latent representations of the intermediate nodes in the metapath-instance connecting the source and destination node, and pass it through a multilayer-perceptron with one neuron at the output layer with a non-linearity $\sigma$ to get the un-normalized attention-weights. Note that, we can use more sophisticated networks to calculate the attention-weights, but use the single layer network (single projection matrix) for simplicity. Note that the representations $h^{0}_{v}$ are the output representations from step 1, i.e., $h^0_v = z_v$.

    \subsubsection{Step 2(b): Aggregating information from different metapaths} We use a simple one-layer multilayer-perceptron to aggregate the information from different metapaths, followed by a non-linearity. This corresponds to projecting each aggregated metapath representation from the step 2(a) and aggregating the projected representations followed by the non-linearity. Formally, we estimate the $l$-layer representation of node $v$ as 
    \begin{equation}
        h^l_v = \sigma\left(\sum_{m \in T}W^l_{m}h^l_{m,v}\right),
    \end{equation}
    where $T$ is the set of metapaths used for the same-type metapath-guided neighbor selection and $W^l_{m}$ is the type-specific projection matrix for $l$th layer and metapath $m$. Similar to the step 2(a), we can use more sophisticated networks for this step too, but use the single layer network (single projection matrix) for simplicity.

\subsection{Putting it together}
Given the two choices we have for aggregating information from the SEN,  we propose two approaches as follows:
\begin{description}[style=unboxed,leftmargin=0cm]
    \item[DHGCN-H: ]Use Hierarchical-Aggregation to aggregate information from the SEN, followed by metapath-aware aggregation from the same-type metapath-guided neighbors.
    \item[DHGCN-S: ]Use Set-Aggregation to aggregate information from the SEN, followed by metapath-aware aggregation from the same-type metapath-guided neighbors.
\end{description}
We call a DHGCN network as $k$-layered, if it aggregates information from same-type metapath-guided neighbors, that are $k$-traversals away. Thus, a 0-layered DHGCN only aggregates information from the SEN, and does not use any information from the same-type metapath-guided neighbors.
\subsection{Training}
Though the proposed approaches can be used for any learning task over graphs (node classification, link-prediction etc.), we present results for the node classification in this paper. As such we minimize the negative log-likelihood between the predicted probabilities and ground truth labels as follows:
\begin{equation}\label{eq:nllh}
    L(\mathcal{G}, F) = -\frac{1}{N}\sum_{i}^{N}\log(F(i, y_i)),
\end{equation}
where $F$ is the model we are training, and $F(i, y_i)$ is the probability predicted for the ground truth class $y_i$ of the training instance $i$.

%% file: methodology.tex
\section{Experimental methodology}\label{sec:experiments}

Though the proposed framework can be used for any learning-on-graph problem, we evaluate it on node classification.  

\subsection{Datasets}
\begin{figure}[!t]%
    \centering
    \subfloat[IMDb]{\includegraphics[width=0.28\columnwidth]{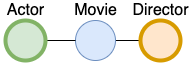} \label{fig:dataset:imdb}}\hfil%
    \subfloat[DBLP]{\includegraphics[width=0.28\columnwidth]{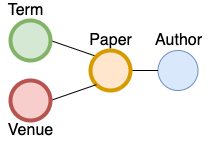} \label{fig:dataset:dblp}}\hfil
    \subfloat[Synthetic]{\includegraphics[width=0.28\columnwidth]{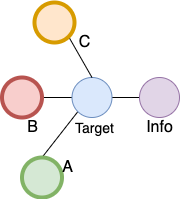} \label{fig:dataset:synthetic}}%
    \caption{Network schema of the datasets. The blue colored circles denote the target node-type. The bold margin circles denote the nodes that appear on the metapaths between the target nodes.}%
    \label{fig:dataset}%
\end{figure}

We use the same two heterogeneous datasets as used in node-classification task in MAGNN~\cite{fu2020magnn}. Specifically, we use IMDb and DBLP datasets.
In addition, we also construct a family of synthetic datasets, with controllable complexity to show the advantages of DHGCN over the competing approaches.
The network schema for each of the datasets is shown in Figure~\ref{fig:dataset}. Table~\ref{tab:datasets} shows the statistics of the real-world datasets. We describe each of these datasets as follows:

\begin{table*}[bt]
\small
\centering
  \caption{Dataset statistics.}
  \label{tab:datasets}
  \begin{tabular}{lcccccc}
  \toprule
           &  No. of &        & No. of &        &        &           \\ 
           &  node   & No. of & edge   & No. of & No. of & No. of    \\ 
  Dataset  &  types  & nodes  & types  & edges  & labels & metapaths \\ 
  \midrule
  IMDb  & $3$ & $11,616$& $2$&$17,106$ & $3$ & $2$\\
  DBLP  & $4$ & $26,128$& $3$&$119,783$ & $4$ & $3$\\
  \bottomrule
  \end{tabular} 
\end{table*}

\begin{description}[style=unboxed,leftmargin=0cm]
\item[IMDb:] IMDb\footnote{\url{https://www.imdb.com/}} 
is an online database about movies and television
programs, including information such as cast, production
crew, and plot summaries. We use the same subset of IMDb as used in~\cite{fu2020magnn}, which contains $4278$ movies, $2081$ directors, and $5257$ actors. The schema for the IMDb dataset is shown in Figure~\ref{fig:dataset:imdb}. The  Movies are labeled as one of three classes (Action, Comedy, and Drama) based on their genre information. The task is to predict the genre of a movie. 
For same-type metapath-guided neighbor selection, we consider the metapaths: Movie-Actor-Movie and Movie-Director-Movie.

\item[DBLP:]DBLP\footnote{\url{https://dblp.org/}} is a computer science bibliography website. We adopt a subset of DBLP extracted by~\cite{ji2010graph, gao2009graph}, containing 4057 authors, 14328 papers, 7723 terms, and 20 publication venues. The schema for the DBLP dataset is shown in Figure~\ref{fig:dataset:dblp}. The authors are divided into four research areas (Database, Data Mining, Artificial Intelligence, and Information Retrieval). The task here is to predict the research area of an author. Similar to IMDb, DBLP is a good dataset for the node classification task, because authors associated with a particular research area tend to publish in particular venues and the features on the paper nodes are also informative. 
For same-type metapath-guided neighbor selection, we consider the metapaths: Author-Paper-Author, Author-Paper-Term-Paper-Author and Author-Paper-Venue-Paper-Author.

\item[Synthetic:] We created a family of synthetic datasets to mimic the schema of real-world applications.  The schema for the synthetic dataset is shown in Figure~\ref{fig:dataset:synthetic}. The task here is to predict the class of the node type `target', where the total number of classes is set to three. The nodes features for the nodes of `target' and `info' are drawn from a multivariate Gaussian with dimensionality $50$. For each class, the coefficients for this mixture follow a multinomial distribution sampled from a Dirichlet distribution. There are three more node types (A, B and C) that we use to create metapaths connecting the `target' node types. We call them bridge node types. The number of nodes of each bridge node type is drawn from a uniform distribution in the range of $[5, 10]$. Each target node is connected to exactly one of the nodes from each of the bridge node type. For each class $c$ and each bridge node type $b$, we have a multinomial distribution sampled from a Dirichlet distribution. For the target nodes belonging to class $c$, we sample the node it connects belonging to the bridge-type $b$ from the corresponding multinomial distribution. To predict the class of a `target' node, we need to estimate the corresponding Gaussian distribution from which the features of the `target' node and the connected `info' nodes are sampled. As such, if sufficient `info' nodes are connected to the target node, the corresponding Gaussian can be estimated. However, if the number of `info' nodes connected to the `target' node is not large enough, we need information from the metapath-neighbors to make accurate predictions. To this extent, we experiment with varying number of `info' nodes connected to the `target' node-type as presented in Section~\ref{sec:results}. For same-type metapath-guided neighbor selection, we consider the metapaths: Target-A-Target, Target-B-Target and Target-C-Target.   
\end{description}
\subsection{Competing approaches}
\begin{description}[style=unboxed,leftmargin=0cm]
\item[RGCN:] Relational Graph Convolution Network (RGCN)~\cite{schlichtkrull2018modeling} extends the idea of GCN for heterogeneous graphs, and uses the type-specific projection matrix for message construction. 
\item[HAN:] Heterogeneous Attention Network (HAN)~\cite{wang2019heterogeneous} estimates metapath-specific node embeddings from different metapath-based homogeneous graphs, and leverages the attention mechanism to
combine them into one vector representation for each node.
\item[MAGNN] For each user-specified meta-path, Metapath Aggregated Graph Neural  Network (MAGNN)~\cite{fu2020magnn}
first performs intra-metapath aggregation by encoding all
the object features along a metapath instance,
and then performs inter-metapath aggregation by attention mechanism.
    
\end{description}
\subsection{Metrics}
For the node classification task, we consider three commonly used metrics: Negative log likelihood, Micro F1 score and  Macro F1 score. 
We discuss each of these metrics below:
\begin{description}[style=unboxed,leftmargin=0cm]
    \item[F1 score:]For a given instance, $\ef$ score is the harmonic mean of the precision and recall based on the predicted classes and the observed classes. It is a popular metric used in binary classification and ranking problems. F1 score reaches its best value at 1 and worst at 0.
    Since, we use multiclass datasets on the node-classification task, we look into two extensions of F1 score for multiclass problems as defined below:\\
    \textit{Micro F1 score: }For the micro F1 setting, we calculate the F1 score by globally counting the total true positives, false negatives and false positives, in order to calculate precision and recall.\\
    \textit{Macro F1 score: }For the macro F1 setting,  we report the mean of $\ef$ score over all the classes.
    \item[Negative log likelihood (NLL):] NLL can be interpreted as the `unhappiness' of the network with respect to its parameters. The higher the NLL, the higher the unhappiness. It is exactly the loss function that we minimize as in Equation~\ref{eq:nllh}.
    
\end{description}
\subsection{Methodology and Parameter selection}
We implemented the proposed framework and the competing approaches in the efficient Deep Graph Learning (DGL)\footnote{\url{https://www.dgl.ai/}}\cite{wang2019dgl} library with PyTorch\footnote{{\url{https://pytorch.org/}}} backend. \\
\textit{Network details: } We used $64$ dimensional representations/embeddings for all the approaches and methods. We train the models with ADAM~\cite{kingma2014adam} optimizer, with the learning rate of $0.001$. For regularization, we used an $L2$ regularization (weight decay) of $0.0001$. In each epoch, we construct mini-batches with neighborhood sampling, and sample at maximum $20$ random neighbors for each relation. The size of a mini-batch is set to $1024$. We perform random walk to sample metapath-guided neighbors for each metapath. For each node and each metapath, we perform $20$ random-walks to sample the metapath-guided neighbors.

For the node classification task, we report results based on a five-fold cross validation. In each fold, we apply early stopping with a patience of five, based on the performance on the NLL metric on the validation set, i.e., the validation loss. We try multiple number of layers (upto 4) for various approaches, and report results for the layer based on the performance on the NLL metric on the validation set.


%% file: results.tex
\section{Results}\label{sec:results}
In this section, we present experiments to demonstrate the accuracy of our approaches for heterogeneous node classification. First, we provide our extensive analysis on the synthetic heterogeneous dataset that we created. After that, we provide and discuss the results for the node classification on the real-world datasets. 

\begin{figure*}[!t]%
    \centering
    \subfloat{\includegraphics[width=0.90\linewidth]{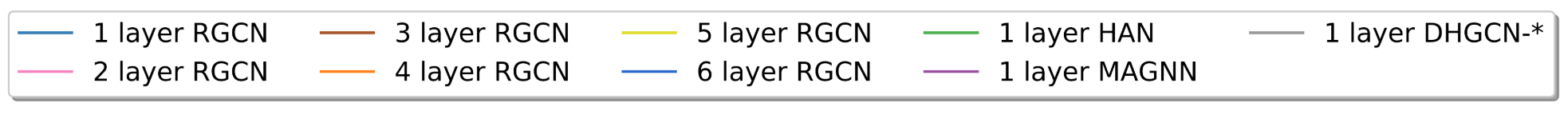} 
     \label{fig:synthetic_results:legend}}%
     \\
    \subfloat[]{\includegraphics[width=0.30\linewidth]{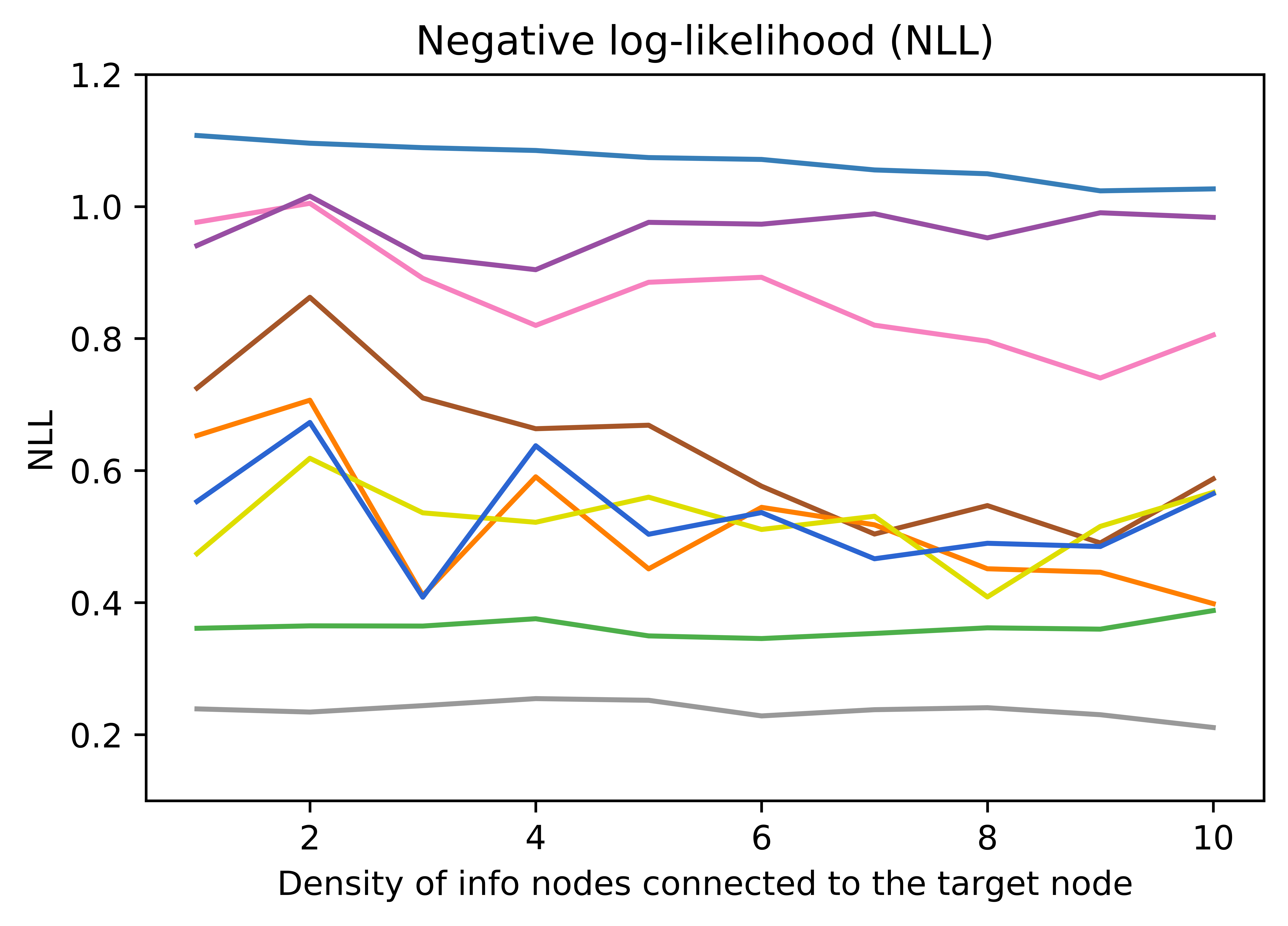}\label{fig:synthetic_results:loss}
    }%
    \subfloat[]{\includegraphics[width=0.30\linewidth]{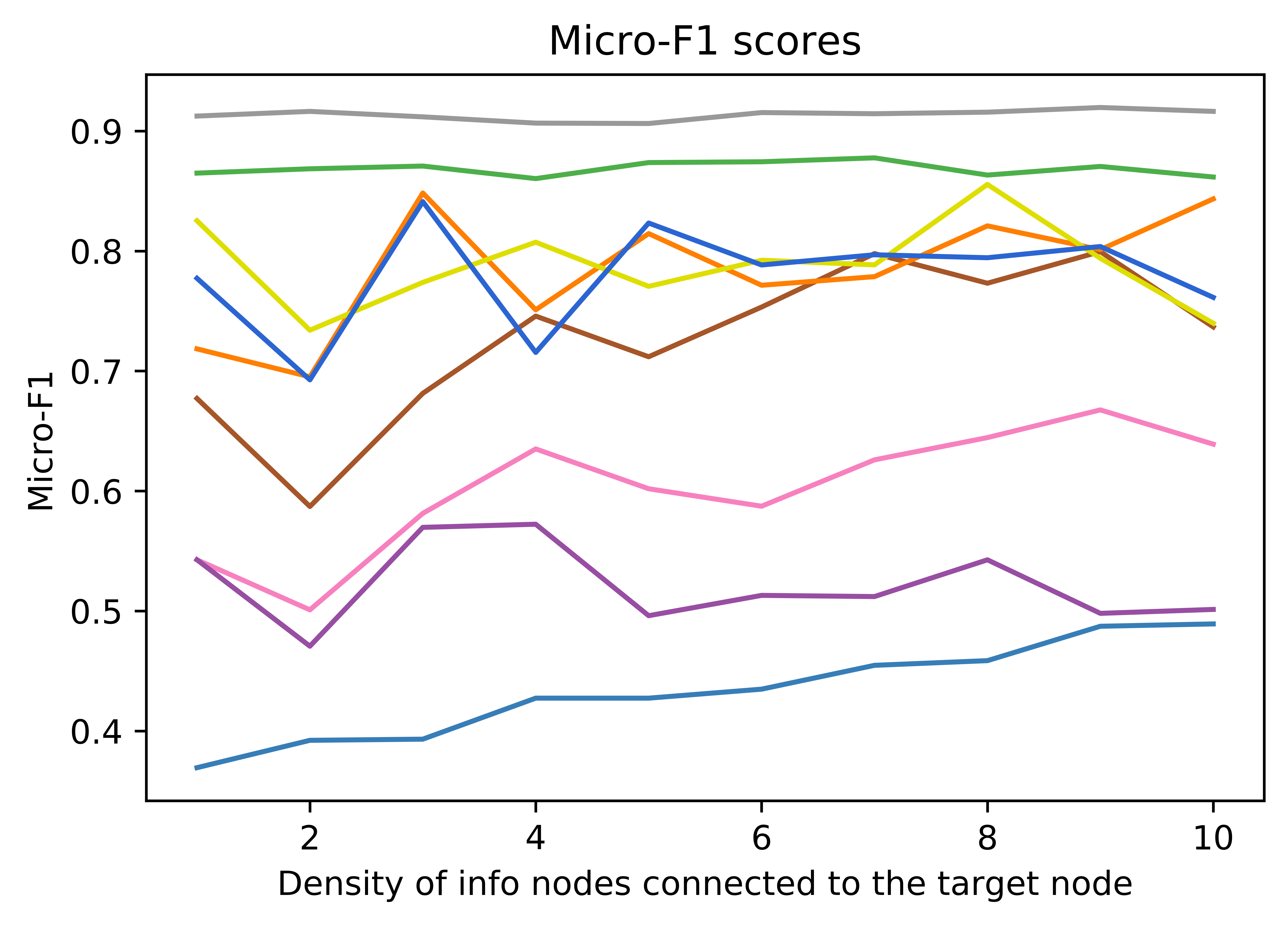}
     \label{fig:synthetic_results:microf1}
     }%
    \subfloat[]{\includegraphics[width=0.30\linewidth]{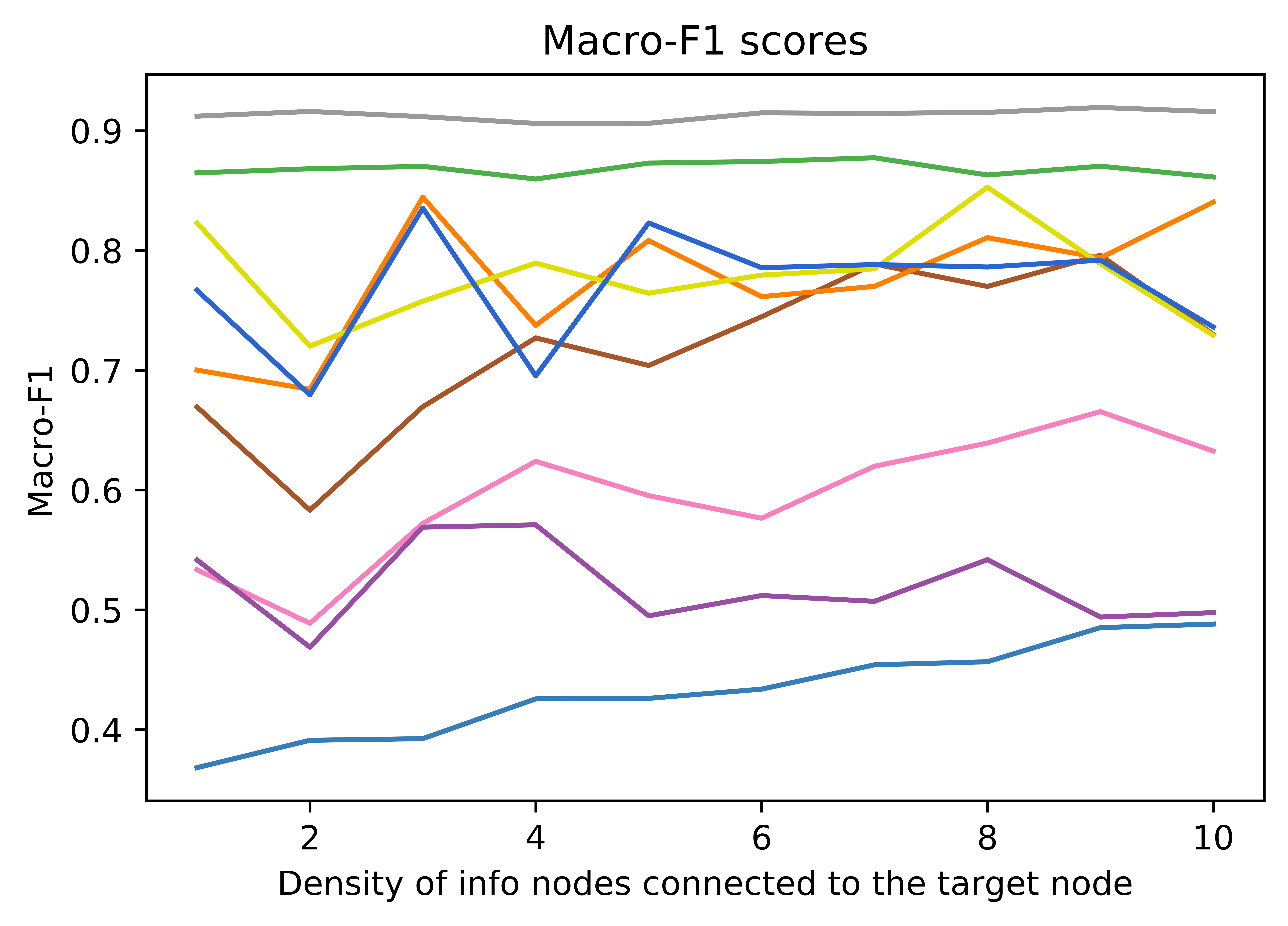} 
     \label{fig:synthetic_results:macrof1}}%
   
    \caption{Performance on the synthetic dataset. }%
    \label{fig:synthetic_results}%
\end{figure*}

\begin{figure*}[!t]%
    \centering
    \subfloat{\includegraphics[width=0.60\linewidth]{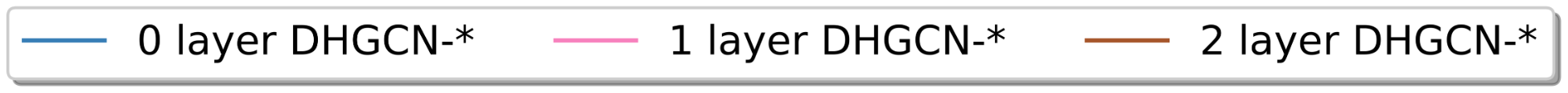} 
     \label{fig:dhgcn_synthetic_results:legend}}%
     \\
    \subfloat[]{\includegraphics[width=0.30\linewidth]{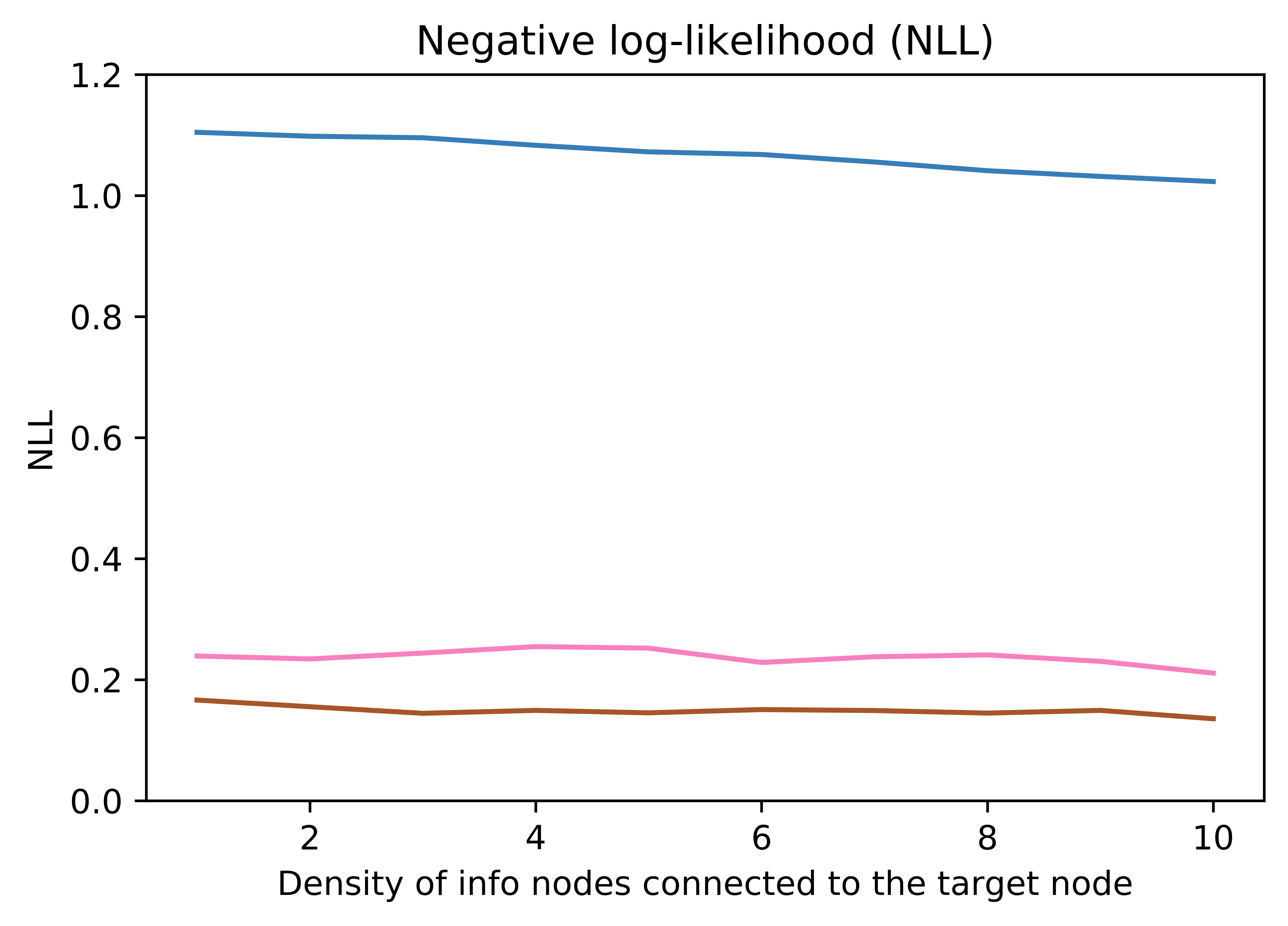}\label{fig:dhgcn_synthetic_results:loss}
    }%
    \subfloat[]{\includegraphics[width=0.30\linewidth]{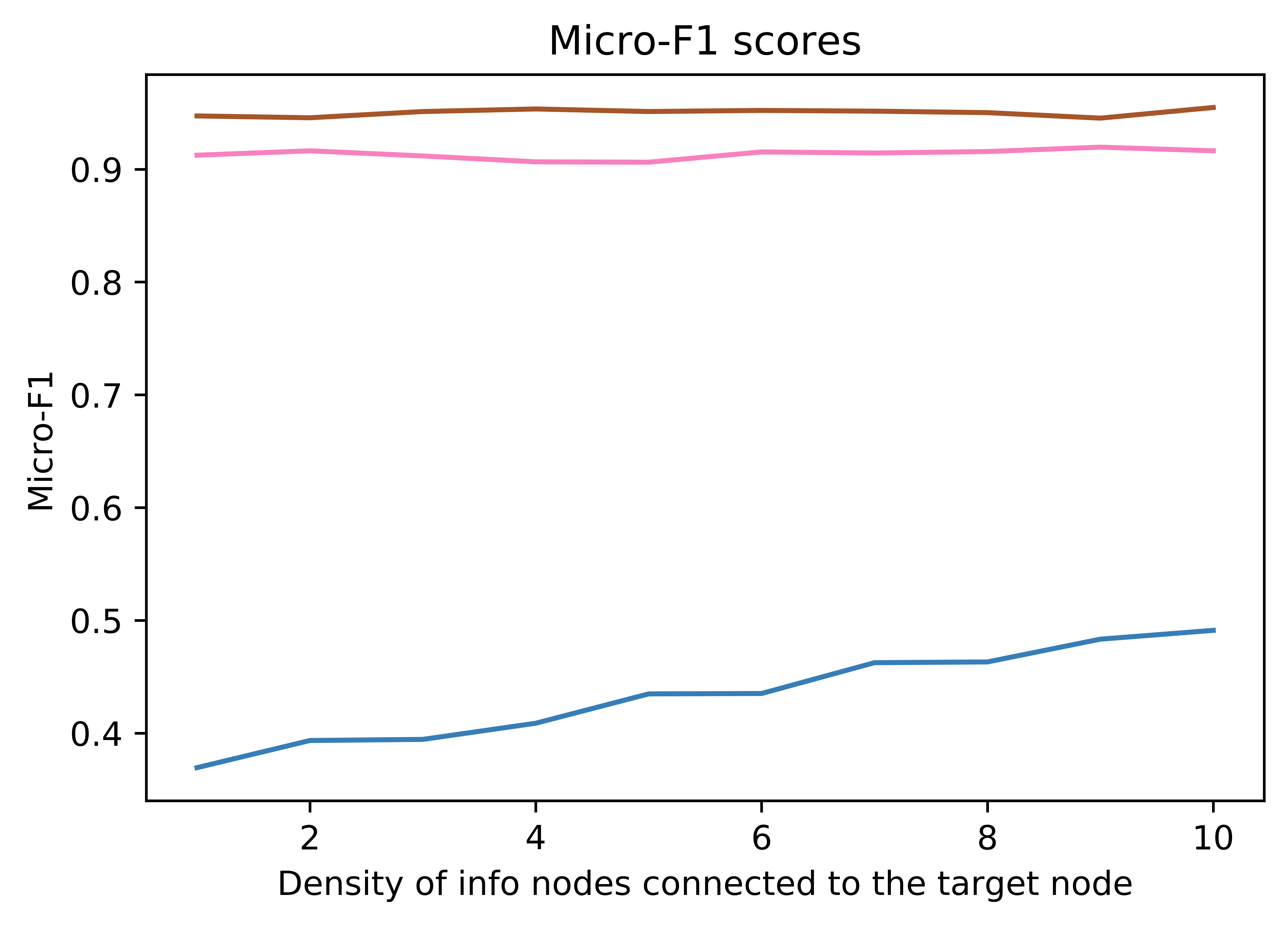}
     \label{fig:dhgcn_synthetic_results:microf1}
     }%
    \subfloat[]{\includegraphics[width=0.30\linewidth]{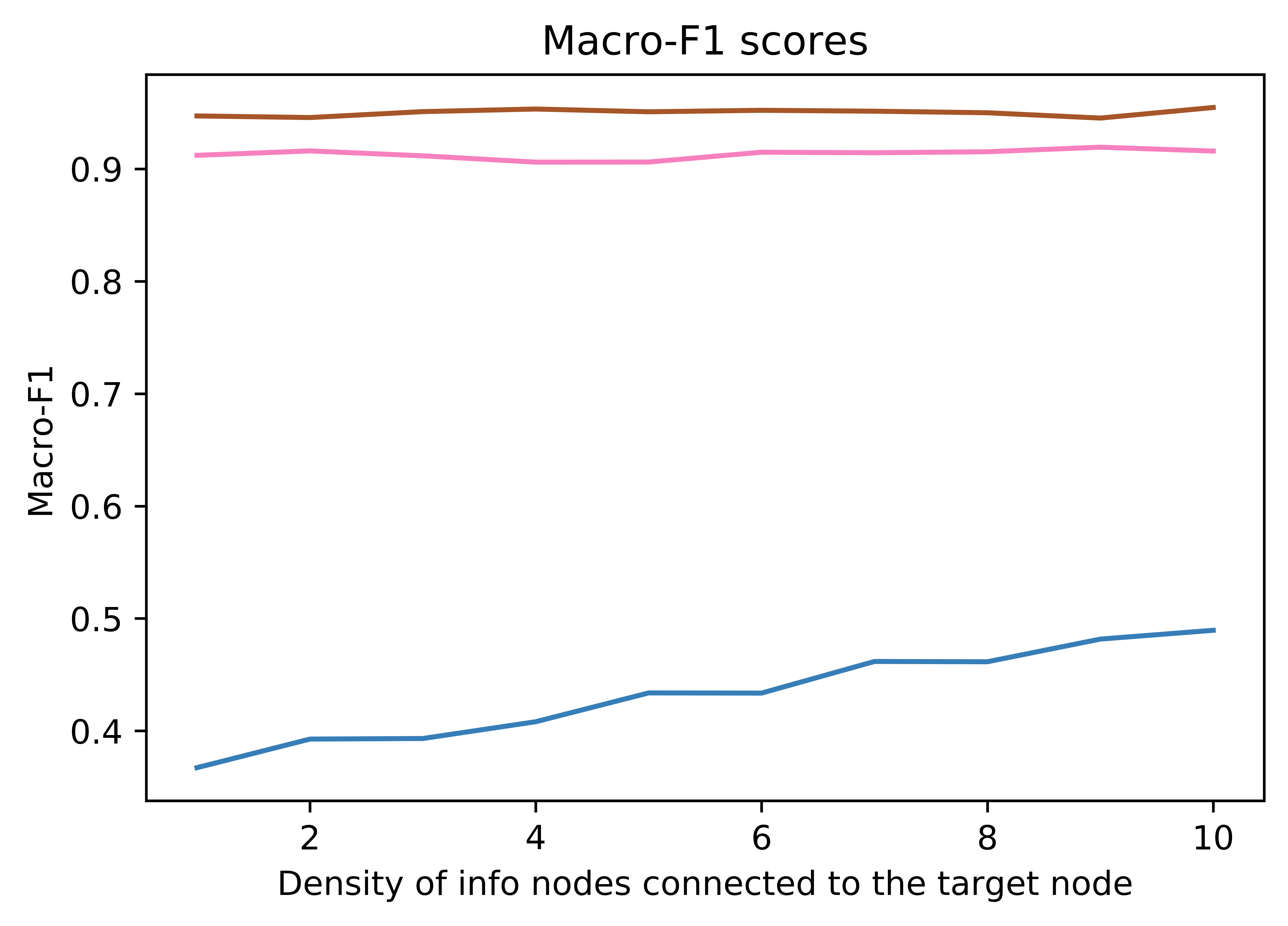} 
     \label{fig:dhgcn_synthetic_results:macrof1}}%
   
    \caption{DHGCN performance variation with number of layers.}%
    \label{fig:dhgcn_synthetic_results}%
\end{figure*}

\subsection{Results on the synthetic dataset: }
We investigate how well different approaches are able to perform when the classification requires information from multiple hops away to make accurate predictions. Specifically, we vary the number of `info' nodes connected to the `target' node, and see if the various approaches are able to get information from the further nodes in order to make accurate predictions. The more `info' nodes connected to the `target' node, the \emph{richer} is the immediate ego-network. Thus, for fewer connected `info' nodes, we need information from the same-type metapath-guided neighbors to make more accurate predictions. We construct $10$ datasets, such that in the $i$th dataset ($1\leq i\leq 10$), for each `target' node, we connect $j$ `info' nodes, where $j$ is sampled from a Gaussian distribution with mean $i$ (with the obvious constraint that $j$ is an integer and $j>0$).

Figures~\ref{fig:synthetic_results:loss}, \ref{fig:synthetic_results:microf1} and~\ref{fig:synthetic_results:macrof1} show the NLL, Micro-F1 and Macro-F1 for various approaches on synthetic datasets with various amount of predictive signals in the immediate ego-net. Note that, both Hierarchical and Set-aggregation are equivalent for the synthetic dataset, so we only show results for the Hierarchical-aggregation based DHGCN. As expected, as the density of the `info' nodes connected to the `target' node increases, the performance of RGGCN and DHGCN approaches improves. By design, MAGNN and HAN do not leverage any information from the `info' nodes, and as such, their performance does not show any trend of change as the number of `info' nodes is varied. Further, we observe that RGCN is unable to leverage information from multiple hops away, and the performance even starts degrading with the number of layers due to the over-smoothing problem \cite{li2018deeper, luan2019break, wu2019simplifying}. Interestingly, HAN yields a significant better result than MAGNN. We conjecture that sequential modeling in the MAGNN results in a loss of information.
In contrast, DHGCN effectively uses the information from multiple hops away and outperforms the competing approaches. We do not see any distinct patterns among the performances of the various DHGCN approaches.

Figure~\ref{fig:dhgcn_synthetic_results} shows the variations of DHGCN as the number of layers increases. DHGCN-* with $0$ layer means we use Hiererchical aggregation to aggregate information over the SEN, but do not use any information from the same-type metapath-guided neighbors. The performance numbers increase along with the number of layers. Clearly, DHGCN based approaches are capable of leveraging from multiple hops away.  Note that, for the synthetic dataset, a one-layer DHGCN actually gets information from three hops away from the underlying graph. A two layer DHGCN is able to get information from five hops away. 

\subsection{Results on the node-classification}

\begin{table}[t!]
\caption{Results on the node classification task: IMDb.}
\label{tab:results:imdb}
\begin{threeparttable}
\begin{tabular}{l c c c}
\toprule
Method     &  NLL & Macro F1 & Micro F1 \\ \midrule
RGCN   & $0.731 \pm 0.013$ & $0.690 \pm 0.009$ & $0.689 \pm 0.010$ \\
MAGNN  & $0.747 \pm 0.016$ & $0.680 \pm 0.005$ & $0.679 \pm 0.006$ \\
HAN    & $0.809 \pm 0.006$ & $0.648 \pm 0.007$ & $0.649 \pm 0.007$ \\[2pt]

DHGCN-S  & $\X{\B{0.729 \pm 0.014}}$ & $\X{\B{0.697 \pm 0.005}}$ & $\X{\B{0.697 \pm 0.006}}$ \\
DHGCN-H  & $\X{\B{0.729 \pm 0.014}}$ & $\X{\B{0.697 \pm 0.005}}$ & $\X{\B{0.697 \pm 0.006}}$ \\
\bottomrule
\end{tabular}
\small
\textbf{Boldfaced} entries correspond to the overall best-performing scheme. \X{Underlined} entries correspond to DHGCN variants whose performance is better than all previously developed methods (RGCN, MAGNN, and HAN).
\end{threeparttable}
\end{table}

\begin{table}[t!]
\centering
  \caption{Results on the node classification task: DBLP.}

\begin{threeparttable}
\begin{tabular}{l c c c}
\toprule
Metric  &  NLL & Macro F1 & Micro F1 \\ \midrule
RGCN    & $0.173 \pm 0.018$ & $0.938 \pm 0.007$ & $0.942 \pm 0.006$ \\
MAGNN   & $0.203 \pm 0.031$ & $0.930 \pm 0.012$ & $0.936 \pm 0.010$ \\
HAN     & $0.260 \pm 0.023$ & $0.918 \pm 0.014$ & $0.924 \pm 0.014$ \\[2pt]

DHGCN-S  & $\X{0.165 \pm 0.019}$ & $\X{0.938 \pm 0.006}$ & $\X{0.942 \pm 0.004}$ \\
DHGCN-H  & $\X{\B{0.155 \pm 0.009}}$ & $\X{\B{0.944 \pm 0.005}}$ & $\X{\B{0.948 \pm 0.005}}$ \\
\bottomrule
\end{tabular}
\small
\textbf{Boldfaced} entries correspond to the overall best-performing scheme. \X{Underlined} entries correspond to DHGCN variants whose performance is better than all previously developed methods (RGCN, MAGNN, and HAN).
\end{threeparttable}
\label{tab:results:dblp}
\end{table}


Table~\ref{tab:results:imdb} shows the performance of different methods on the IMDb dataset. All the DHGCN approaches outperform the competing approaches achieving up to $0.3\%$, $1.0\%$ and $1.2\%$ improvement over the next best approach, which is RGCN for all three metrics.
Note that both Hierarchical-Aggregation and Set-Aggregation are equivalent in the case of IMDb, as there is only one level of information in the IMDb schema. Further, RGCN outperforms HAN and MAGNN.

Table~\ref{tab:results:dblp} shows the performance of the methods on the DBLP dataset. In the same fashion, DHGCN approaches, in general, outperform all of the competing approaches on the DBLP dataset on the NLL metric, achieving up to $10.4\%$ improvement over the next best performing baseline (RGCN).
On the Macro F1 and Micro F1, though DHGCN-H outperforms the competing baselines, while DHGCN-S performs at par with RGCN. Specifically, DHGCN approaches outperform RGCN by achieving up to $0.6\%$ on both the Macro F1 and Micro F1 metrics. Similar to IMDb, RGCN and DHGCN approaches outperform HAN and MAGNN.



%% file: conclusion.tex
\section{Conclusion}\label{sec:conclusion}
In this paper, we proposed solutions to two major limitations of GCN approaches for heterogeneous graphs: (i) over-smoothing, leading to low expressive power as a result of shallow models; or (ii) inability to explore the complete ego-network, leading to missing out some nodes, which might provide important signals for the task at hand. To this extent, our solution \emph{Deep Heterogeneous Graph Convolutional Network} (DHGCN), is a purposefully designed class of GCN models for heterogeneous graphs. DHGCN uses a two-step schema-aware hierarchical approach. In the first step, DHGCN aggregates information from the immediate ego-network that is derived from the schema. In the second step, DHGCN aggregates information from the same-type metapath-guided neighbors.
On a variety of synthetic and real-world datasets, we show that the proposed approaches outperform the competing baselines.

There are many avenues of optimization of our approaches worth investigating, for example, using stronger models to estimate the attention weights for Step-2 aggregation, such as deeper neural networks, using multi-head attention etc.  It is also important to perform further benchmarking on heterogeneous datasets generated from more complex schemes. This constitutes our future work.